%% file: main.tex
\documentclass[letterpaper,10pt,conference]{ieeeconf}

\IEEEoverridecommandlockouts
\usepackage[T1]{fontenc}
\usepackage{times}
\usepackage{graphicx}
\usepackage{amsmath}
\usepackage{amssymb}

\usepackage{amsthm}
\usepackage{cite}
\usepackage{xcolor}

\theoremstyle{plain}
\newtheorem{theorem}{Theorem}[section]

\theoremstyle{definition}

\newtheorem{example}[theorem]{Example}

\newtheorem{remark}[theorem]{Remark}

\usepackage{algorithm}
\usepackage{float} %
\usepackage{booktabs}
\usepackage{algpseudocode}
\usepackage{colortbl}
\usepackage{capt-of}
\makeatletter
\let\NAT@parse\undefined
\makeatother
\usepackage[hidelinks]{hyperref}

\title{\LARGE \bfseries Zero-Shot Reactive Obstacle Avoidance for Generative Robot Policies}

\author{Weihang Guo and Lydia E.\ Kavraki%
\thanks{The authors are with the Department of Computer Science, Rice University,
Houston, TX, USA. {\tt\small \{wg25,kavraki\}@rice.edu}. LEK is also affiliated with the Ken Kennedy Institute at Rice University.
}
}

\begin{document}
\maketitle
\thispagestyle{empty}
\pagestyle{empty}

\input{content/01_abstract}

\input{content/02_intro}
\input{content/03_related}

\input{content/04_problem_definition}
\input{content/04_methods}
\input{content/05_experiments}

\input{content/07_conclusion}

\section*{Acknowledgements}
LEK and WG have been supported in part by NSF 2411219 and Rice University Funds.

\bibliographystyle{IEEEtran}
\bibliography{ref}

\end{document}

%% file: content/01_abstract.tex
\begin{abstract}
    We propose \underline{NUDGE}~(\underline{N}udge \underline{U}pdate via
    \underline{D}ifferentiable \underline{GE}ometry), a training-free
    obstacle-avoidance procedure that can be incorporated in any robot
    policy based on diffusion or flow matching, including diffusion policies
    and vision-language-action models.
    Our work injects gradients from a signed distance field, a function
    returning each point's distance to the nearest obstacle, into the policy
    at inference time to steer it away from obstacles. It supports any
    common action parameterization, from absolute or relative joint poses to
    end-effector poses, through a differentiable joint-trajectory decoder.
    Experiments show that NUDGE preserves the policy's task distribution and
    runs reactively in real time.
\end{abstract}

%% file: content/02_intro.tex
\section{Introduction}
\begin{figure*}[ht]
    \centering
    \includegraphics[width=0.85\linewidth]{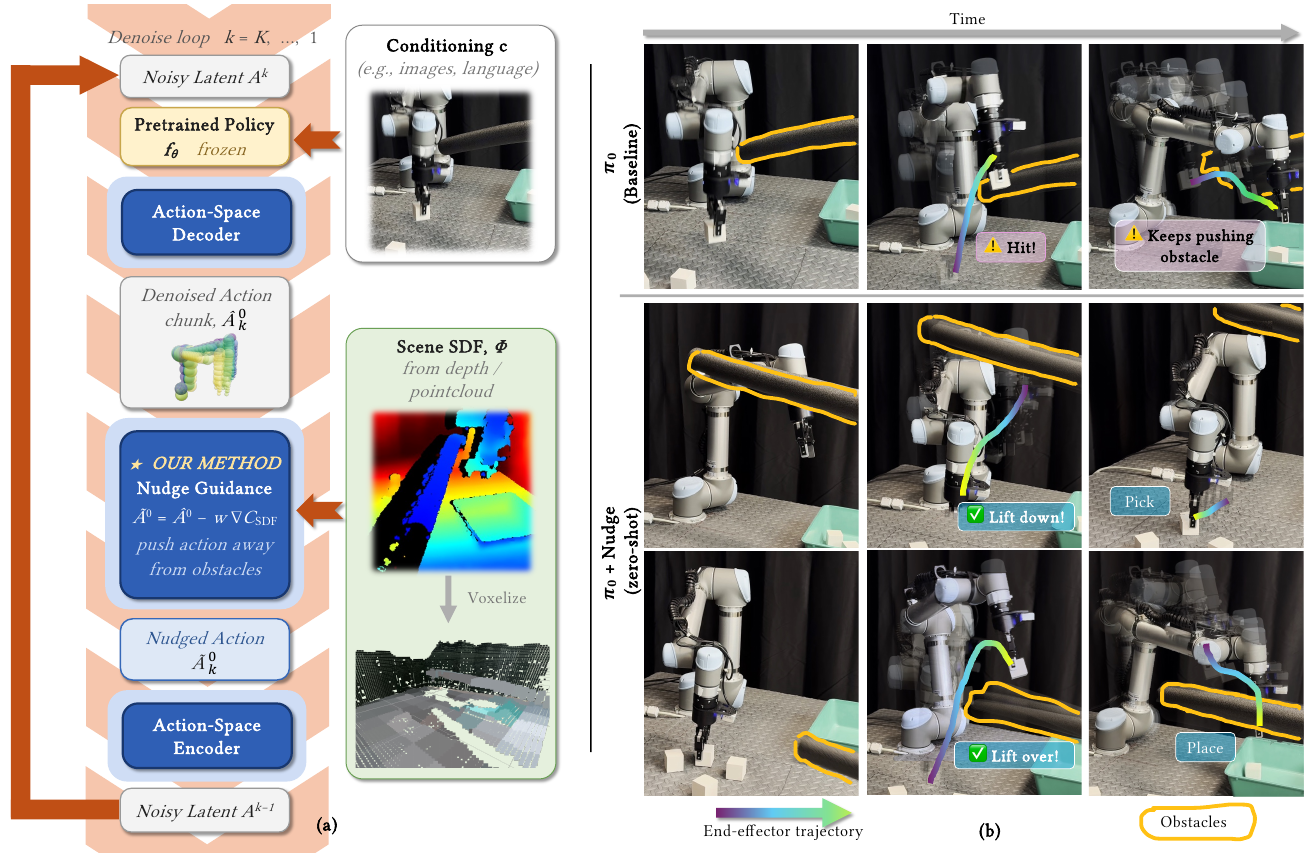}
    \caption{\textbf{NUDGE provides zero-shot collision avoidance for
    generative robot policies at inference time.}
    (a) From a live point cloud, NUDGE computes a signed distance field in \emph{real time} to react to scene changes, injects its gradient into the policy's denoising loop, and steers the
    predicted action chunk away from obstacles.
    (b) We fine-tuned a $\pi_{0}$ policy for a table-cleaning task, where the policy 
    learns to model the contact and task distribution from demonstrations. \textbf{Top:} the unguided baseline collides with the pool noodle.
    \textbf{Middle \& Bottom:} NUDGE avoids the obstacle at the extra cost of only
    2.5~ms per action chunk.}
    \label{fig:fig1}
\end{figure*}

Modern robotics policies~\cite{chi2024diffusionpolicy, black2024pi0, black2025pi} learn closed-loop visuomotor control and handle implicit task constraints that are hard to
specify analytically but are easy to demonstrate, such as keeping a cup upright while pouring,
maintaining contact while wiping, or holding a grasp through transport. However, policies
are trained on obstacle-free demonstrations. When deployed in environments with novel obstacles, the robot may collide with them~\cite{jung2025rail}. This problem raises safety concerns. 

We consider inference-time obstacle avoidance for a pretrained action-chunk
policy that was trained only on obstacle-free demonstrations. We propose
NUDGE~(\underline{N}udge \underline{U}pdate via \underline{D}ifferentiable
\underline{GE}ometry), a training-free procedure that adds obstacle
avoidance without overriding the learned task distribution. NUDGE requires
only that the policy exposes a denoised action chunk at each sampling step,
a property shared by diffusion, flow-matching, and vision-language-action
samplers. At each step, NUDGE
evaluates the predicted action chunk using a signed distance field~(SDF), which maps each workspace point to its signed distance from the nearest obstacle surface, and nudges the chunk
against the cost gradient before the sampler re-encodes it to the next noise
level. Subsequent sampling steps absorb the nudge, so the policy's output
stays close to the demonstration manifold while avoiding obstacles encoded
in the SDF.

In this paper, we contribute 1) A training-free, model-agnostic guidance procedure for obstacle
    avoidance that works on any action-chunk policy based on diffusion or
    flow matching, covering diffusion policies and vision-language-action
    models.
2) Empirical evidence on two simulation benchmarks that NUDGE
    preserves the policy's learned task distribution while avoiding obstacles.
3) Deployment on a fine-tuned $\pi_{0}$ VLA driving a UR5 against
    dynamic obstacles, with NUDGE adding only 2.5\,ms per action chunk.

%% file: content/03_related.tex
\section{Related Work}

We distinguish two largely separate threads of work that use generative
models for robot motion. \emph{Motion planners} take a start and goal
configuration and output an entire collision-free trajectory in a single
pass, with obstacle geometry encoded as a cost or constraint~\cite{janner22planning, carvalho2025mpd, saha2024edmp, seo2025presto, yang2025deep, xiao2023safediffuser, zhang2026constrained, romer2025diffusion, yang2026safeflowmatcher}.
\emph{Task policies} instead learn
closed-loop visuomotor behavior from demonstrations and produce short
action chunks conditioned on observations. Task policies encode task-level
constraints such as grasp orientation or surface contact implicitly through
training data, without a dedicated geometry-aware collision cost.
\emph{NUDGE targets the task policies category}, adding inference-time
collision avoidance to pretrained task policies without disturbing the
learned task distribution.

\subsection{Generative Task Policies}
\label{sec:task-policies}

Task policies map current observations to short action chunks, learning
task behavior from demonstrations. Action chunking was introduced by
ACT~\cite{zhao2023act}, and Diffusion Policy~\cite{chi2024diffusionpolicy}
formulates it as conditional denoising over chunks. $\pi_{0.5}$~\cite{black2025pi} and RDT~\cite{liu2025rdt} extend the same idea to
vision-language-action models by pairing a pretrained VLM with a
flow-matching or diffusion action head. NUDGE also supports other newer
flow-matching policies such as Streaming Flow Policy~\cite{jiang2025sfp},
Action-to-Action Flow Matching~\cite{jia2026action}, and
VITA~\cite{gao2026vita}. Both classes learn task-level
constraints~(distribution) implicitly through demonstration data, such as keeping a cup
upright or maintaining contact with a surface. Because obstacle avoidance is
not part of their training signal, the policy can cause collisions between the robot and the environment
when obstacles are present. NUDGE adds the missing obstacle awareness
at inference without retraining or modifying the policy. We note that autoregressive based policies
such as OpenVLA~\cite{kim2025openvla} and RT-2~\cite{zitkovich2023rt} fall outside NUDGE scope. They generate actions one
token at a time, with no intermediate estimate of the full chunk for
guidance to act on.

\subsection{Guided Diffusion and Obstacle Avoidance}

Guidance steers pretrained diffusion models toward auxiliary objectives
without retraining, by adding a cost gradient to the learned
score~\cite{dhariwal2021diffusion,bansal2023universal,chung2023diffusion} or
by hard-projecting onto a constraint set~\cite{wang2023ddnm}. NUDGE follows
Diffusion Posterior Sampling~\cite{chung2023diffusion} in evaluating the cost
on the predicted action chunk rather than the noisy intermediate, and
propagates the gradient analytically through differentiable signed distance field and robot kinematics.

Within robotics research, DynaGuide~\cite{du2025dynaguide},
VLS~\cite{liu2026vls}, ACG~\cite{park2025acg}, and Lan-o3dp~\cite{li2025lano3dp} apply similar
inference-time guidance using learned dynamics, VLM-synthesized rewards, or
object-centric representations, but none target whole-body collision
avoidance. RAIL~\cite{jung2025rail} is a reachability-based safety filter that runs on
top of a pretrained Diffusion Policy: each predicted chunk is checked for
collision, and an unsafe chunk is replaced with a precomputed safe backup.
NUDGE instead edits the chunk in place.
Classical filters such as CBFs~\cite{ames2019cbf}, reachability analysis~\cite{bansal2017hamilton},
and residual policies~\cite{johannink2019residual} apply orthogonally to our
inference-time setting, as does safe-RL exemplified by
CPO~\cite{achiam2017cpo}. Dedicated reactive planners such as cuRoboV2~\cite{sundaralingam2026curobov2}
SDFs~\cite{millane2024nvblox}, STORM~\cite{bhardwaj2022storm},
Neural~MP~\cite{dalal2025neuralmp}, and Deep Reactive
Policy~\cite{yang2025deep} produce obstacle-aware trajectories but lack the
task semantics encoded by demonstration training.
Concurrent with our work, OmniGuide~\cite{song2026omniguide} also guides generalist policies
at inference time that guides a sparse set of Cartesian probe points, primarily the
end-effector. NUDGE instead focuses on whole-body obstacle avoidance, supporting joint- and
end-effector-space actions via a differentiable decoder.

%% file: content/04_methods.tex
\section{Method}
\label{sec:method}

We consider inference-time obstacle avoidance for a pretrained action-chunk
policy that was trained only on obstacle-free demonstrations. The pretrained
policy may be any action-chunk policy based on diffusion or flow matching,
including diffusion policies and vision-language-action models. At
deployment, we are given the current robot state $q_t$, the conditioning input
$\mathbf{c}$ that the policy expects, and a signed distance field $\Phi$ of
the workspace constructed in \emph{real time} to react to scene changes. 
The goal is to produce action chunks that complete the task
while minimizing the collision cost induced by $\Phi$, without retraining the
policy and without pushing its outputs off the learned task distribution.

At a high level, NUDGE treats each intermediate denoised action chunk as a
candidate short-horizon robot motion. A differentiable decoder maps the
chunk from the policy's action representation to a sequence of future joint
configurations. NUDGE then evaluates whole-body obstacle clearance along
this trajectory and differentiates the resulting collision cost
with respect to the action chunk. The chunk is shifted along the negative
cost gradient and passed back to the sampler. Thus, subsequent steps
incorporate the correction and retain the task behavior encoded by the
pretrained policy.

\subsection{Iterative Action-Chunk Generative Policies}\label{sec:action_chunk}

Modern visuomotor policies~\cite{chi2024diffusionpolicy, black2025pi} model
$p(\mathbf{A}\mid\mathbf{c})$ over action chunks
$\mathbf{A}=[\mathbf{a}_1,\ldots,\mathbf{a}_H]\in\mathbb{R}^{H\times d}$ of
joint- or end-effector~(EEF)-space actions. Conditioning $\mathbf{c}$ may combine images,
language, and proprioception. Sampling refines Gaussian noise
$\mathbf{A}^K\sim\mathcal{N}(0,I)$ through
$\mathbf{A}^K\to\cdots\to\mathbf{A}^0$ over $K$ steps. At step $k$, the
network $f_\theta(\mathbf{A}^k,k,\mathbf{c})$ predicts noise, velocity, or a clean target;
a formulation-specific decoder $h$ converts it into a clean-chunk estimate
$\hat{\mathbf{A}}^0_k$. NUDGE requires this intermediate estimate:
$
  \mathbf{A}^k \;\longmapsto\;
  \hat{\mathbf{A}}^0_k = h\bigl(\mathbf{A}^k, f_\theta(\mathbf{A}^k, k, \mathbf{c})\bigr)
  \;\longmapsto\; \mathbf{A}^{k-1},
  \label{eq:abstract_step}
$
where guidance modifies $\hat{\mathbf{A}}^0_k$ between clean-chunk prediction and
re-encoding to the next noise or time level.

\begin{example}[DDIM diffusion policy]
For a DDPM-trained diffusion policy~\cite{ho2020ddpm} sampled with
DDIM~\cite{song2021ddim}, $f_\theta = \boldsymbol{\epsilon}_\theta$ is a noise
predictor and the denoised action chunk is
$
  \hat{\mathbf{A}}^0_k = \frac{1}{\sqrt{\bar{\alpha}_k}}
  \Bigl( \mathbf{A}^k - \sqrt{1 - \bar{\alpha}_k}\,
  \boldsymbol{\epsilon}_\theta(\mathbf{A}^k, k, \mathbf{c}) \Bigr),
$
followed by re-diffusion
$\mathbf{A}^{k-1} = \sqrt{\bar{\alpha}_{k-1}}\, \hat{\mathbf{A}}^0_k +
\sqrt{1 - \bar{\alpha}_{k-1}}\, \boldsymbol{\epsilon}_\theta(\mathbf{A}^k, k, \mathbf{c})$.
\end{example}

\begin{example}[Flow-matching policy]
For a flow-matching policy~\cite{black2025pi} along the linear path
$\mathbf{A}^k = (1 - t_k)\, \mathbf{A}^0 + t_k\, \boldsymbol{\epsilon}$ with
$t_k \in [0, 1]$, $f_\theta = v_\theta$ is a velocity field and the
denoised action chunk is 
$
  \hat{\mathbf{A}}^0_k = \mathbf{A}^k - t_k\, v_\theta(\mathbf{A}^k, t_k, \mathbf{c}),
$
followed by an Euler integration step toward $t_{k-1}$.
\end{example}

\begin{remark}[Generality]
The interface is independent of network architecture and conditioning
modality, and supports absolute or relative joint and EEF poses.
\end{remark}

\subsection{Action-Space Decoder}\label{sec:action_decoder}

To remain agnostic to the policy's action representation, we evaluate
all downstream costs on a predicted joint trajectory $q_{t+1:t+H}$ obtained
from the action chunk through a differentiable decoder. Let
$q_t \in \mathbb{R}^{n_q}$ be the current joint state for an $n_q$-DOF arm,
and let $\mathcal{D}$ be a differentiable decoder
\begin{equation}
  q_{t+i} = \mathcal{D}_i(\mathbf{A}; q_t), \qquad i = 1, \ldots, H,
  \label{eq:decoder}
\end{equation}
that maps the action chunk and the current joint state to the $i$-th future
joint configuration. Table~\ref{tab:decoders} lists $\mathcal{D}$ for the
action representations in common use, covering both joint-space and
end-effector-space variants.

\begin{table}[h]
\centering
\caption{Action-space decoders}
\label{tab:decoders}
\resizebox{\linewidth}{!}{%
\begin{tabular}{lll}
\toprule
Action representation & Notation & Decoder $q_{t+i} = \mathcal{D}_i(\mathbf{A}; q_t)$ \\
\midrule
Joint pose   & $\mathbf{a}_i = q^{\mathrm{tgt}}_{i}$ & $q_{t+i} = \mathbf{a}_i$ \\
Joint delta  & $\mathbf{a}_i = \Delta q_i$           & $q_{t+i} = q_t + \sum_{j=1}^{i} \mathbf{a}_j$ \\
EEF pose     & $\mathbf{a}_i = T^{ee}_i$             & $q_{t+i} = \mathrm{IK}(\mathbf{a}_i;\, q_{t+i-1})$ \\
EEF delta    & $\mathbf{a}_i = \Delta T^{ee}_i$      & $q_{t+i} = \mathrm{IK}\!\bigl(T^{ee}_t \cdot \prod_{j=1}^{i} \mathbf{a}_j;\, q_{t+i-1}\bigr)$ \\
\bottomrule
\end{tabular}
}
\end{table}

\subsection{SDF-Guided Denoising}\label{sec:sdf}

We adopt a classifier-guidance~\cite{dhariwal2021diffusion} formulation of
obstacle avoidance. The cost is evaluated on the predicted joint trajectory
$q_{t+i} = \mathcal{D}_i(\mathbf{A}; q_t)$, $i = 1, \ldots, H$, with
$\mathcal{D}$ drawn from Table~\ref{tab:decoders} to match the policy's
action representation.
We assume access to an SDF updated from the latest scene observation,
$\Phi: \mathbb{R}^3 \to \mathbb{R}$ of the obstacles in the workspace, with
$\Phi(p) > 0$ in free space, $\Phi(p) < 0$ inside an obstacle, and
$|\Phi(p)|$ equal to the Euclidean distance from $p$ to the nearest obstacle
surface. For each predicted configuration we query $\Phi$ at points on the
robot's collision geometry. Following common practice in SDF-based motion
planning~\cite{ratliff2009chomp, sundaralingam2023curobo}, let
$\mathcal{S}_l = \{ (c^{l,j}, r^{l,j}) \}_{j=1}^{N_l}$ be the sphere decomposition
of link $l$ (positions and radii in link frame) and let
$T_l(q) \in \text{SE}(3)$ be the forward-kinematics transform of link $l$ at
configuration $q$. We define the per-chunk collision cost as a one-sided hinge
over all robot spheres at all chunk steps:
\begin{equation}
  \mathcal{C}_{\mathrm{SDF}}(\hat{\mathbf{A}}^0) =
  \sum_{i=1}^{H} \sum_{l} \sum_{j=1}^{N_l}
  \bigl[\, d_{\mathrm{safe}} - \Phi\!\bigl( T_l(q_{t+i})\, c^{l,j} \bigr) + r^{l,j} \,\bigr]_{+}^{2},
  \label{eq:cost}
\end{equation}
where $[\cdot]_{+} = \max(0, \cdot)$ and $d_{\mathrm{safe}} \geq 0$ is a safety
margin. The cost is zero whenever every sphere clears every obstacle by at least
$d_{\mathrm{safe}}$. Held objects enter the same sum through collision spheres
attached to the gripper. We evaluate the discrete configurations in the
action chunk; the objective does not check continuous swept volumes between them.

The gradient of $\mathcal{C}_{\mathrm{SDF}}$ with respect to joint
configurations is computed by automatic differentiation through the forward
kinematics:
$  
  \frac{\partial \mathcal{C}_{\mathrm{SDF}}}{\partial q_{t+i}} =
  \sum_{l, j} 2 \bigl[d_{\mathrm{safe}} - \Phi + r^{l,j}\bigr]_{+}
  \bigl( -\nabla_{\!p} \Phi \bigr)^{\!\top}
  J^{l,j}(q_{t+i}),
$
where $J^{l,j}(q) = \partial (T_l(q)\, c^{l,j}) / \partial q$ is the standard
manipulator Jacobian of sphere $j$ on link $l$. Mapping the gradient back to
action space is a single chain-rule application through $\mathcal{D}$:
\begin{equation}
  \frac{\partial \mathcal{C}_{\mathrm{SDF}}}{\partial \mathbf{a}_m} =
  \sum_{i=1}^{H}
  \left( \frac{\partial \mathcal{D}_i}{\partial \mathbf{a}_m} \right)^{\!\!\top}
  \frac{\partial \mathcal{C}_{\mathrm{SDF}}}{\partial q_{t+i}},
  \qquad m = 1, \ldots, H.
  \label{eq:chainrule}
\end{equation}

\begin{example}[Joint-delta decoder]
For the joint-delta row of Table~\ref{tab:decoders}, \textit{i.e.}, 
$q_{t+i} = q_t + \sum_{j=1}^{i} \Delta q_j$, the Jacobian $\partial \mathcal{D}_i / \partial \Delta q_m$ is the identity when $m \leq i$ and zero otherwise, so Eq.~\eqref{eq:chainrule} collapses to a reverse cumulative sum: $\frac{\partial \mathcal{C}_{\mathrm{SDF}}}{\partial \Delta q_m} =
  \sum_{i=m}^{H} \frac{\partial \mathcal{C}_{\mathrm{SDF}}}{\partial q_{t+i}},$ for $m = 1, \ldots, H$.
\end{example}

\subsection{Guided Sampling Step}\label{sec:nudge_guide}

At each sampling step we inject the SDF gradient by nudging the denoised
action chunk $\hat{\mathbf{A}}^0_k$ before the sampler re-encodes it to the next
level:
\begin{equation}
  \tilde{\mathbf{A}}^0_k = \hat{\mathbf{A}}^0_k - w_k \, \nabla_{\!\hat{\mathbf{A}}^0} \mathcal{C}_{\mathrm{SDF}}(\hat{\mathbf{A}}^0_k),
  \label{eq:shift}
\end{equation}
where $w_k \geq 0$ sets the guidance strength; zero disables guidance.
The sampler re-encodes $\tilde{\mathbf{A}}^0_k$ instead of
$\hat{\mathbf{A}}^0_k$ to produce $\mathbf{A}^{k-1}$, as illustrated below.

\begin{example}[Guided DDIM step]\label{exa:ddim_guided}
For a DDIM sampler, $\tilde{\mathbf{A}}^0_k$ replaces $\hat{\mathbf{A}}^0_k$ in
the re-diffusion update:
$
  \mathbf{A}^{k-1} = \sqrt{\bar{\alpha}_{k-1}}\, \tilde{\mathbf{A}}^0_k +
  \sqrt{1 - \bar{\alpha}_{k-1}}\, \boldsymbol{\epsilon}_\theta(\mathbf{A}^k, k, \mathbf{c}).
$
\end{example}

\begin{example}[Guided flow-matching step]\label{exa:flow_guided}
For a flow-matching sampler with the linear path,
the Euler step toward $t_{k-1}$ can be written as a linear interpolation
between the current latent and the denoised action chunk; replacing
$\hat{\mathbf{A}}^0_k$ with $\tilde{\mathbf{A}}^0_k$ gives the guided update
$
  \mathbf{A}^{k-1} = \frac{t_{k-1}}{t_k}\, \mathbf{A}^k +
  \frac{t_k - t_{k-1}}{t_k}\, \tilde{\mathbf{A}}^0_k.
$
\end{example}

At deployment, each control step executes the first
$H_{\mathrm{exec}} \leq H$ predicted actions and re-runs the sampler against
the latest world-frame SDF $\Phi$. Algorithm~\ref{alg:guided} gives the DDIM
procedure. Flow matching replaces the update
with Example~\ref{exa:flow_guided}.
Raw scales depend on the policy's joint- or EEF-action coordinates. For
example, LIBERO~\cite{liu2023libero}'s action-to-displacement conversion introduces an
approximately $0.0085$\,m factor into the decoder Jacobian, so a large
numerical $w_k$ need not imply a large physical correction. We can
reparameterize the scale as
\begin{equation}
  \rho_k = \frac{w_k\kappa}{1+w_k\kappa}\in[0,1),
  \qquad w_k = \frac{\rho_k}{\kappa(1-\rho_k)},
  \label{eq:normalized_scale}
\end{equation}
where $\kappa>0$ is the median nonzero $\ell_\infty$ gradient norm on a fixed
calibration set in normalized action coordinates. It measures the typical
gradient magnitude, allowing search over bounded $\rho_k$. This invertible
mapping preserves the update and subsequent clipping.

\begin{algorithm}
\caption{Guided Action Chunk Generation}
\label{alg:guided}
\begin{algorithmic}[1]
\Require Joints $q_t$, conditioning $\mathbf{c}$, policy $f_\theta$,
         denoising decoder $h$, action decoder $\mathcal{D}$, SDF $\Phi$,
         horizon $H$, steps $K$, scales $\{\rho_k\}_{k=1}^{K}$,
         calibration factor $\kappa>0$
\State Sample $\mathbf{A}^{K} \sim \mathcal{N}(0, I)$
\For{$k = K, K-1, \ldots, 1$}
  \State $w_k \gets \rho_k/[\kappa(1-\rho_k)]$ \hfill{\small\emph{// Eq.~\eqref{eq:normalized_scale}}}
  \State $\hat{\mathbf{A}}^0_k \gets h\bigl(\mathbf{A}^k,\, f_\theta(\mathbf{A}^k, k, \mathbf{c})\bigr)$ \hfill{\small\emph{// decode, Sec.~\ref{sec:action_chunk}}}
  \State $g \gets \nabla_{\!\hat{\mathbf{A}}^0} \mathcal{C}_{\mathrm{SDF}}(\hat{\mathbf{A}}^0_k)$ \hfill{\small\emph{// Eqs.~\eqref{eq:decoder},~\eqref{eq:cost},~\eqref{eq:chainrule}}}
  \State $\tilde{\mathbf{A}}^0_k \gets \hat{\mathbf{A}}^0_k - w_k\, g$ \hfill{\small\emph{// Eq.~\eqref{eq:shift}}}
  \State $\mathbf{A}^{k-1} \gets \mathrm{ReEncode}_k\bigl(\mathbf{A}^k,\, \tilde{\mathbf{A}}^0_k\bigr)$ \hfill{\small\emph{// Ex.~\ref{exa:ddim_guided}/\ref{exa:flow_guided}}}
\EndFor
\State \Return $\mathbf{A}^0$
\end{algorithmic}
\end{algorithm}

%% file: content/05_experiments.tex
\begin{table*}[t]
\centering
\small
\caption{Sphere-manifold results}
\label{tab:sphere}
\begin{tabular}{lcccccc}
\toprule
& & Goal & \multicolumn{2}{c}{Collision} & \multicolumn{2}{c}{Manifold} \\
\cmidrule(lr){3-3} \cmidrule(lr){4-5} \cmidrule(lr){6-7}
Method & Success $\uparrow$ & Reach $\uparrow$ & Coll.\ Ep.\ $\downarrow$ & Coll.\ Rate $\downarrow$ & On-Sphere $\uparrow$ & Sphere Err.\ (cm) $\downarrow$ \\
\midrule
\rowcolor{black!8}
Vanilla & 37.0\% & 82.0\% & 54.0\% & 16.4\% & 84.3\% & 2.78 \\
Post-Projection & 25.0\% & 29.0\% & 14.5\% & 4.7\% & 56.4\% & 4.84 \\
OmniGuide~\cite{song2026omniguide} & 39.0\% & 81.0\% & 52.0\% & 15.98\% & 84.9\% & 2.71 \\
RAIL~\cite{jung2025rail} & 37.5\% & 37.5\% & 0.0\% & 0.00\% & 82.9\% & 2.73 \\\midrule
NUDGE, $\rho_k=0.01$ & 42.0\% & 73.0\% & 47.0\% & 13.8\% & 80.5\% & 3.00 \\
NUDGE, $\rho_k=0.04$ & \underline{66.5\%} & 73.0\% & 10.0\% & 0.91\% & 83.6\% & 2.86 \\
NUDGE, $\rho_k=0.07$ & \textbf{71.5\%} & 73.5\% & 2.5\% & 0.21\% & 83.0\% & 2.87 \\
\bottomrule
\\
\end{tabular}

\textit{Note:} \emph{Success} is the integrated
metric: an episode succeeds only when the robot reaches the goal
\emph{and} is collision-free throughout. The remaining columns decompose
performance into three groups that a useful safe controller must balance: \emph{Goal} (Reach, the fraction of
episodes that reach the goal regardless of collisions), \emph{Collision}
(Coll.~Ep., the fraction of episodes with any collision; Coll.~Rate, the
fraction of control steps across all episodes during which the robot is in
collision), and \emph{Manifold} (On-Sphere, the percentage
of trajectory steps within $5$\,cm of the sphere surface; Sphere~Err.,
the mean distance from the sphere surface across all steps).
\end{table*}
\section{Experiments}

Our experiments answer three questions:
\begin{itemize}
    \item \emph{Does inference-time SDF guidance prevent collisions while
    preserving the learned task manifold?}
    Section~\ref{exp:sphere} answers this on a controllable sphere-constraint
    task whose manifold is explicit and measurable.
    \item \emph{Does NUDGE generalize to policies whose task constraints are
    only implicit in demonstration data, without hurting task performance?}
    Section~\ref{exp:dp_sim} answers this by inserting unseen obstacles into
    LIBERO-Object scenes.
    \item \emph{Does NUDGE work on a real-robot VLA, and is SDF guidance a
    real-time bottleneck?} Section~\ref{exp:vlash} answers this by deploying
    NUDGE on a fine-tuned $\pi_0$ driving a UR5 against dynamic obstacles,
    with an end-to-end timing breakdown.
\end{itemize}

\subsection{Baselines and Robot Collision Geometry}
We include RAIL~\cite{jung2025rail} and OmniGuide~\cite{song2026omniguide} as our baselines. RAIL enforces collision avoidance as a hard constraint through post-generation verification. After the policy has generated a complete action chunk, RAIL checks the chunk for collisions. If it is unsafe, RAIL replaces it with a safe stopping trajectory instead of guiding or modifying the action chunk. By rejecting rather than correcting unsafe trajectories, RAIL may sacrifice task success to enforce hard safety.

\begin{figure}[h]
\centering
\includegraphics[width=\linewidth]{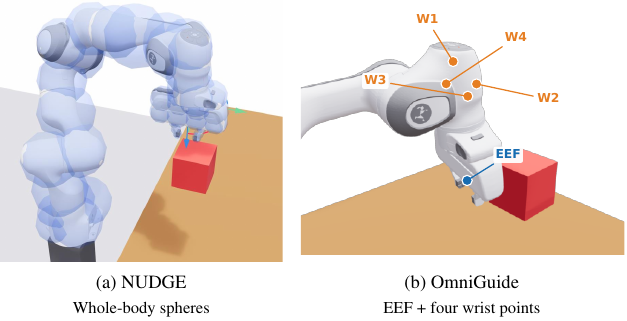}
\caption{Collision-query geometry: NUDGE's whole-body spheres versus OmniGuide's EEF point and four wrist probe.}
\label{fig:sdf_ill}
\end{figure}

OmniGuide~\cite{song2026omniguide}, concurrent work only on arXiv, guides pretrained policies at
inference time, but evaluates collision costs using only a sparse set of
Cartesian probe points around the end effector and wrist, as shown in
Fig.~\ref{fig:sdf_ill}. OmniGuide constructs the evironment with VGGT~\cite{wang2025vggt}. Consequently, it does not explicitly account for
collisions involving the remaining robot links. By contrast, we approximate
robot links with spheres for \emph{whole-body} collision checking based on cuRoboV2~\cite{sundaralingam2026curobov2}. Each SDF query, minus the sphere radius, gives the clearance
used in $\mathcal{C}_{\mathrm{SDF}}$.

\subsection{Sphere Constraint}
\label{exp:sphere}

\begin{figure}[h]
\centering
\begin{minipage}[c]{0.35\linewidth}
    \includegraphics[width=\linewidth]{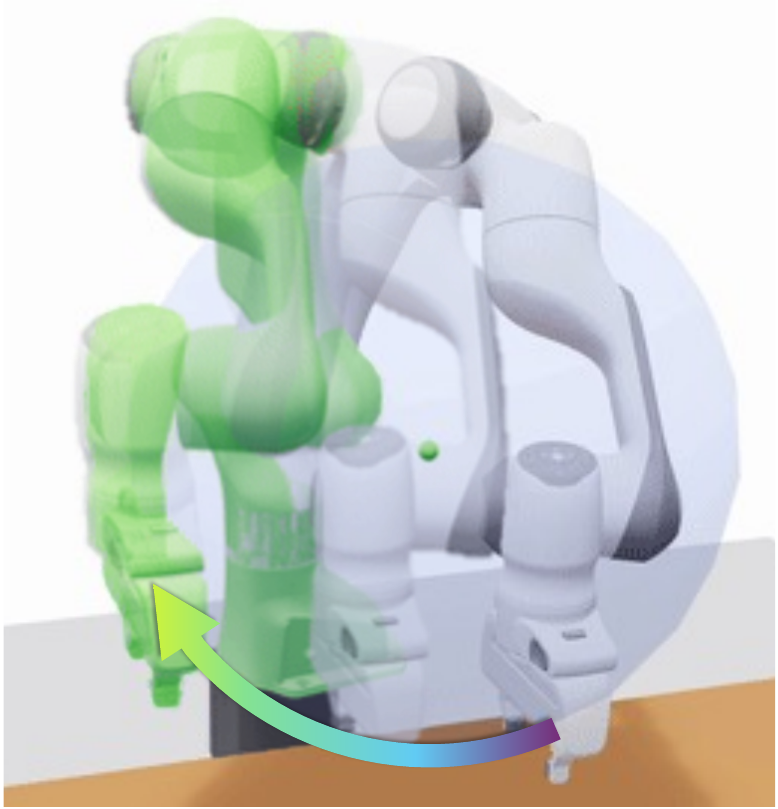}
\end{minipage}%
\begin{minipage}[c]{0.43\linewidth}
    \includegraphics[width=\linewidth]{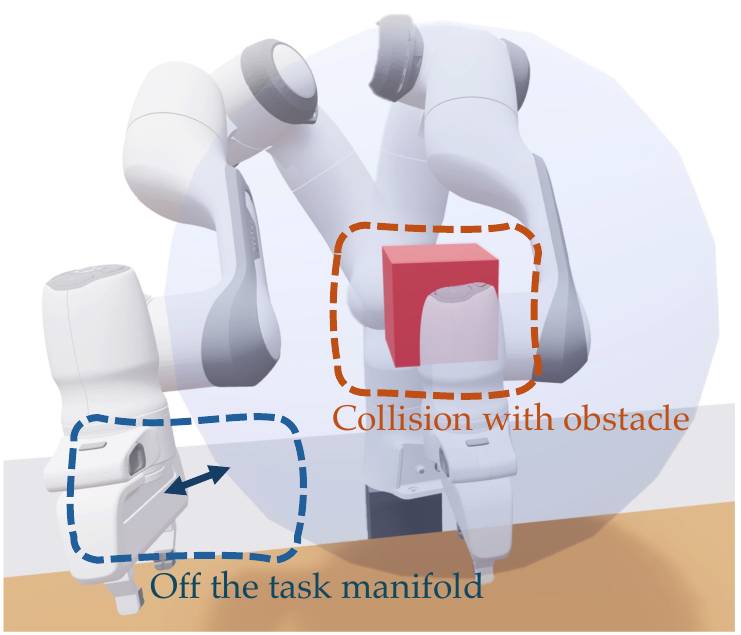}
\end{minipage}%
\caption{\textbf{Left.} The robot trajectory from start to goal with end-effort constrained on the sphere manifold. \textbf{Right.} Failure modes on the sphere
    constraint task: collides with the obstacle and pushes the end-effector off the learned task manifold.}
\label{fig:sphere_failure}
\end{figure}

The sphere-constrained task isolates manifold preservation using an
explicit, quantifiable manifold. The sphere serves as a
proxy for harder-to-measure manifolds in grasping, pushing, or balancing,
and we measure manifold violation as
$e = \bigl|\,\|\mathbf{p}_\text{eef} - \mathbf{c}\| - r\,\bigr|$, the
distance from the end-effector to the sphere surface
(Fig.~\ref{fig:sphere_failure}). This setup enables us to quantifiably measure how close the algorithm maintains the task manifold constraint.  

A \texttt{Vanilla} diffusion policy~\cite{chi2024diffusionpolicy} is
trained on $21{,}760$ obstacle-free trajectories of a 7-DOF
Franka Panda on a sphere of radius $r=0.25$\,m, conditioned on current joints
and goal pose. Action chunks are $H=16$ joint deltas sampled with DDIM over
$K=10$ steps. Test boxes have side lengths $6$--$12$\,cm and centers within
$\pm8$\,cm of the sphere center; start and goal are at least $60^\circ$ apart
on the surface. We evaluate $200$ episodes with identical starts, goals,
and obstacles across methods. We compare
against diffusion policy~(\texttt{Vanilla}), \texttt{Post-Projection},
which applies joint-space collision correction after sampling,
and \texttt{RAIL}~\cite{jung2025rail}, which substitutes an
ARMTD~\cite{holmes2020armtd} failsafe whenever the predicted chunk could
collide. NUDGE uses the three $\rho_k$ settings in Table~\ref{tab:sphere},
fixed across denoising steps, with replanning
every eight control steps.

All methods in Table~\ref{tab:sphere} share the pretrained policy.
Reach requires joint distance to the goal below $0.3$\,rad. Success
additionally requires no collision. Coll.~Ep. and Coll.~Rate measures the percentages of episodes and
steps with contact. On-Sphere measures steps within 5\,cm of the manifold;
Sphere Err. is the mean distance to it.

NUDGE achieves the highest overall success rate by balancing manifold
preservation and obstacle avoidance, where each baseline collapses on one
or more axes. In both \texttt{Post-Projection} and \texttt{RAIL}~\cite{jung2025rail}, 
the correction is decoupled from the policy's
score and can disagree with the policy's intended direction, causing the
robot to oscillate and sometimes settle in a local minimum. NUDGE applies
the gradient inside the policy's own denoising loop, so the correction
stays aligned with the learned task distribution while pushing the chunk
away from collision.

\subsection{Collision Avoidance in LIBERO and Latency}
\label{exp:dp_sim}
\label{exp:omniguide}

\begin{figure}[!ht]
\centering
\includegraphics[width=0.8\linewidth]{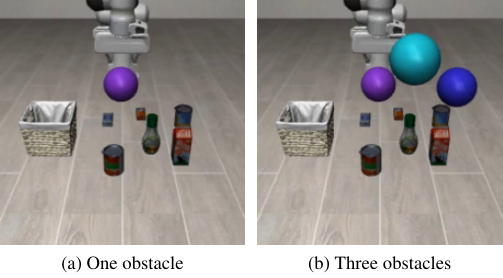}
\caption{One- and three-obstacle LIBERO-Object scenes with matched initial robot/object poses and camera view. We note that the additional obstacles may make a collision-free grasp infeasible. }
\label{fig:libero_obstacles}
\end{figure}

We next test whether NUDGE generalizes when the task constraint is implicit
and not directly measurable, the typical setting for learned manipulation. We use the fine-tuned
$\pi_{0.5}$~\cite{black2025pi} checkpoint, a flow-matching VLA with
EEF-delta actions.
LIBERO-Object~\cite{liu2023libero} comprises $10$ pick-and-place tasks
with obstacle-free demonstrations. The fine-tuned $\pi_{0.5}$ achieves
$98.4\%$ obstacle-free success. Action chunks are $H=10$ EEF deltas sampled
with the flow-matching Euler integrator over $K=10$ steps. We inject a spherical obstacle between the object and the basket,
lifted 18\,cm, with radius $6$--$8$\,cm, $\pm1.5$\,cm horizontal and
$\pm1$\,cm vertical jitter, and one of six colors. The three-obstacle
condition retains this sphere and adds two $6$--$8$\,cm spheres along the robot's critical path
(Fig.~\ref{fig:libero_obstacles}).
Both unguided $\pi_{0.5}$, OmniGuide,
and NUDGE are evaluated on 100 episodes per method and condition, with ten
initial states per task and identical obstacle draws across methods.

\begin{table}[h]
\centering
\caption{LIBERO-Object results}
\label{tab:libero_obs_blocking}
\small
\setlength{\tabcolsep}{3pt}
\begin{tabular}{lrrrrrrrr}
\toprule
 & \multicolumn{2}{c}{Unguided}
 & \multicolumn{2}{c}{OmniGuide}
 & \multicolumn{2}{c}{RAIL}
 & \multicolumn{2}{c}{NUDGE} \\
\cmidrule(lr){2-3}
\cmidrule(lr){4-5}
\cmidrule(lr){6-7}
\cmidrule(lr){8-9}
Task
 & SR $\uparrow$ & CR $\downarrow$
 & SR $\uparrow$ & CR $\downarrow$
 & SR $\uparrow$ & CR $\downarrow$
 & SR $\uparrow$ & CR $\downarrow$ \\
\midrule
\multicolumn{9}{l}{\textit{One obstacle}} \\
\midrule
Can      & 50 & 5.70  & 60 & 15.19 & 10 & 0.00 & 100 & 0.00 \\
Cheese   & 40 & 6.56  & 40 & 17.25 & 0  & 0.00 & 60  & 9.71 \\
Dressing & 20 & 4.55  & 40 & 0.09  & 0  & 0.00 & 40  & 0.00 \\
BBQ      & 0  & 4.72  & 10 & 2.22  & 0  & 0.00 & 10  & 0.03 \\
Ketchup  & 20 & 11.78 & 0  & 10.55 & 0  & 0.00 & 0   & 0.10 \\
Tomato   & 0  & 31.24 & 0  & 28.21 & 0  & 0.00 & 0   & 10.66 \\
Butter   & 90 & 6.14  & 80 & 5.89  & 0  & 0.00 & 90  & 0.00 \\
Milk     & 0  & 23.14 & 0  & 19.86 & 0  & 0.00 & 30  & 2.03 \\
Pudding  & 70 & 28.58 & 70 & 25.37 & 0  & 0.00 & 80  & 0.50 \\
Juice    & 10 & 9.05  & 0  & 0.59  & 0  & 0.00 & 30  & 0.00 \\
\midrule
Mean     & 30 & 13.15 & 30 & 12.52 & 1  & 0.00 & 44  & 2.30 \\
\midrule
\midrule
\multicolumn{9}{l}{\textit{Three obstacles}} \\
\midrule
Can      & 10 & 25.61 & 10 & 23.22 & 0 & 0.00 & 20 & 4.62 \\
Cheese   & 0  & 21.31 & 10 & 17.43 & 0 & 0.00 & 40 & 23.56 \\
Dressing & 10 & 27.16 & 0  & 40.97 & 0 & 0.00 & 0  & 15.45 \\
BBQ      & 0  & 40.41 & 10 & 31.57 & 0 & 0.00 & 0  & 2.76 \\
Ketchup  & 0  & 32.48 & 0  & 31.45 & 0 & 0.00 & 0  & 6.86 \\
Tomato   & 0  & 49.41 & 0  & 44.17 & 0 & 0.00 & 0  & 24.90 \\
Butter   & 10 & 31.26 & 30 & 15.15 & 0 & 0.00 & 40 & 1.83 \\
Milk     & 0  & 42.48 & 0  & 37.28 & 0 & 0.00 & 0  & 10.90 \\
Pudding  & 10 & 45.22 & 10 & 33.93 & 0 & 0.00 & 30 & 12.26 \\
Juice    & 0  & 67.28 & 0  & 60.76 & 0 & 0.00 & 0  & 36.93 \\
\midrule
Mean     & 4  & 38.26 & 7  & 33.59 & 0 & 0.00 & 13 & 14.01 \\
\bottomrule
\\
\end{tabular}
\textit{Note:} SR denotes task success rate. CR denotes the percentage of recorded control steps with robot-obstacle contact.
All values are percentages.

\end{table}

All three methods use the same $\pi_{0.5}$ checkpoint;
NUDGE and OmniGuide additionally share the EEF-action decoder.
NUDGE constructs a cuRoboV2 ESDF from a single metric-depth view with 2\,cm voxels.
Simulator masks exclude the robot, target, basket, and table.
Its robot representation comprises 41 Panda spheres
(Fig.~\ref{fig:sdf_ill}) and a 7\,cm payload sphere at a fixed
offset from \texttt{panda\_hand}.
We set $\rho_k=0.85$ (Eq.~\eqref{eq:normalized_scale})
across all ten tasks.
OmniGuide uses VGGT~\cite{wang2025vggt} to reconstruct geometry
from front, overhead, and agent RGB views, with object-aware CLIP
for semantic target filtering.
The collision geometries are five points near the end-effector~(Fig.~\ref{fig:sdf_ill}).
Following a hyperparameter search, we use a guidance scale of $0.4$
and EEF/wrist thresholds of $0.2/0.1$\,m.
For RAIL, the released implementation does not provide a perception
pipeline for constructing its geometric representation from our
LIBERO observations. We therefore use simulator ground-truth
geometry for collision checking and backup planning.

In RAIL, 113 of the 200 LIBERO episodes
terminate after exhausting the backup-search budget, and another
76 reach the episode time limit.
Backup control accounts for 76.3\% and 83.1\% of executed steps
in the one- and three-obstacle conditions, respectively.
The position-based recovery objective does not encode the complete
grasp-and-place sequence. This suggests that frequent intervention can limit task progress even when robot-obstacle contact is avoided.

We measure the environment-representation latency of NUDGE and
OmniGuide on 25 paired frames from five LIBERO rollouts.
Each pipeline runs independently on an RTX~4090 after five
warmup requests.
Timing includes input preprocessing, representation construction,
and the service round trip, while excluding rendering, disk I/O,
policy inference, and action execution.
Mean latency is 907.44\,ms for OmniGuide and 3.32\,ms for NUDGE,
a ratio of approximately \textbf{273.3$\times$ faster}. This enables NUDGE to deliver real-time, reactive performance in the real world, which we will introduce next. 

\subsection{Real-Robot Evaluation and System Latency}
\label{exp:vlash}
\label{exp:perception_latency}

\begin{figure}[h]
    \centering
    \includegraphics[width=\linewidth]{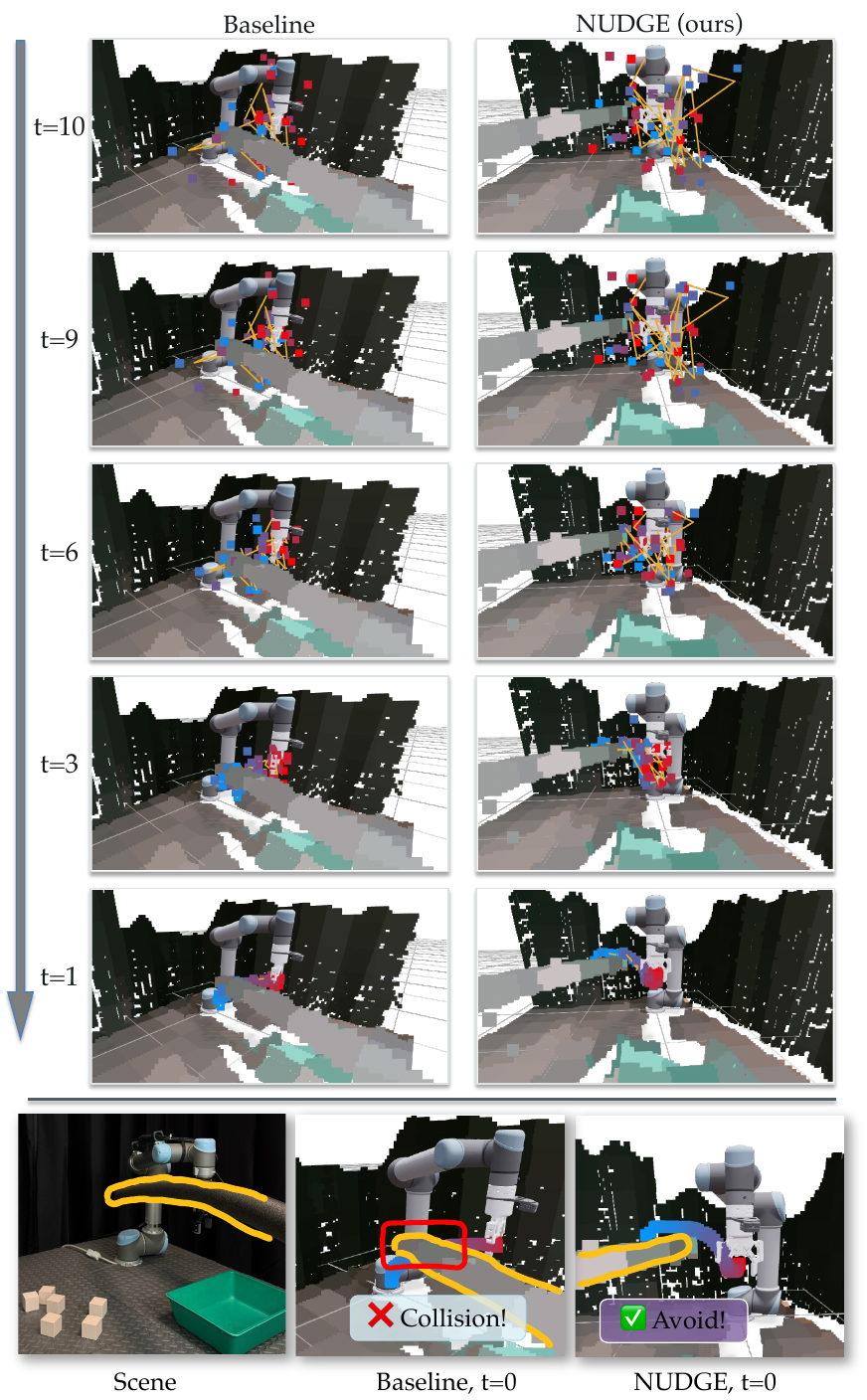}
\caption{\textbf{Top:} Denoising visualization for a single action chunk.
    The policy predicts a $50$-step chunk of joint deltas. The actions are mapped to end-effector pose and rendered in a color spectrum
    across denoising steps, from random noise at $t = 10$ down to the final
    denoised chunk at $t = 0$. The environment is shown as a voxel map.
    \textbf{Bottom, left:} Real-world setup; the obstacle is a pool noodle.
    \textbf{Bottom, middle and right:} Final action chunk at $t = 0$ for the
    unguided baseline and NUDGE, respectively. The baseline is unaware of
    the obstacle and collides with it. NUDGE pushes the chunk away from the
    obstacle.}
    \label{fig:real_exp}
\end{figure}

\begin{table}[h]
\centering
\caption{UR5 inference latency (ms)}
\label{tab:vlash-breakdown}
\small
\setlength{\tabcolsep}{3pt}
\begin{tabular}{lrrr}
\toprule
Stage & $\pi_0$ & $+$ NUDGE & $\Delta$ \\
\midrule
Observation I/O and transfer &   5.06 &   5.54 & $+0.48$ \\
\midrule
Vision/language encoding                   &   8.94 &   9.01 & $+0.07$ \\
Causal prefill                             &  17.02 &  17.48 & $+0.46$ \\
Denoising loop (10 steps)                  &  72.40 &  76.38 & $+3.98$ \\
\quad Action-expert step, mean              &   7.13 &   7.20 & $+0.07$ \\
\rowcolor[gray]{0.9} \quad Guidance (last 2 steps)      &   ---  &   3.76 & $+3.76$ \\
Post-processing                            &   0.16 &   0.20 & $+0.04$ \\
\midrule
Total, eager & 104.76 & 109.90 & $+5.14$ \\
Total, compiled & \phantom{0}34.41 & \phantom{0}36.87 & $+2.46$ \\
\bottomrule
\end{tabular}
\end{table}

\begin{figure}[h]
\centering
\includegraphics[width=\linewidth]{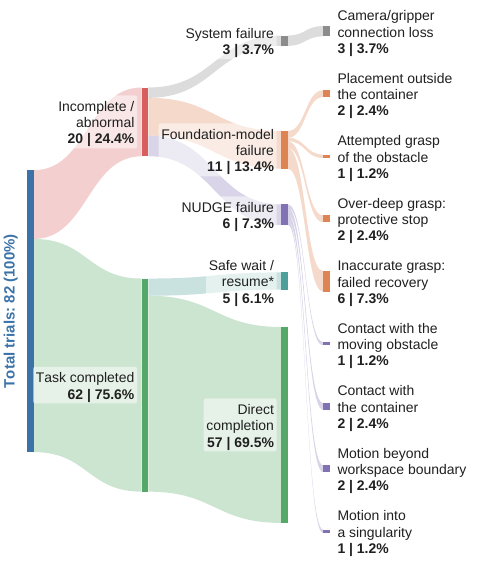}
\caption{\textbf{Real UR5 outcome breakdown} of all 82 consecutive. In each trial, the robot picked a cube on the table and placed it into a container while a foam obstacle moved continuously through the pick-and-place workspace. 
    NUDGE success rate was 76.5\%. The foundation model, rather than NUDGE, was the primary bottleneck, accounting for 11 of the 19 incomplete trials. (*) denotes autonomous
hold-and-resume during temporary blockage, without task assistance.}
\label{fig:real_robot_outcomes}
\end{figure}

We deploy a fine-tuned $\pi_{0}$~\cite{black2024pi0} on a UR5. In each trial,
the robot picks one of six cubes from a table and places it in a container.
An experimenter continuously moves one foam obstacle through the corridor
between the cubes and container, challenging both reaching and transport.
We use an RTX 5090, CUDA 12.8, PyTorch 2.11, and $\pi_0$ in bf16.
A cuRoboV2~\cite{sundaralingam2026curobov2} node fuses live depth
from a single RGB-D camera into a 2\,cm voxel map. Joint state, base RGB,
and wrist RGB arrive over ZMQ.
The base policy $\pi_0$ is fine-tuned on language-conditioned pick-and-place,
with chunk size $H=50$ and $K=10$ flow-matching steps. SDF guidance is
injected only at the last two denoising steps. This late-step guidance is also used in image
generation~\cite{bansal2023universal}.

Across 82 consecutive trials, NUDGE completes 62 (75.6\%) without obstacle
or container contact. These six trials account
for the 7.3\% failures attributed to NUDGE. Of the remaining 14 failures,
11 concern grasping or placement and three are camera/gripper connection losses.
In comparison, across 18 trials, unguided $\pi_0$ completes three (16.7\%),
collides in 13 (72.2\%), and has two grasping failures.
NUDGE achieves \textbf{4.5 times higher completion rate} than baseline. The 82-trial outcome
decomposition appears in Fig.~\ref{fig:real_robot_outcomes}.
Figure~\ref{fig:real_exp} visualizes the guided denoising process.
On the RTX~5090, Table~\ref{tab:vlash-breakdown} compares $\pi_0$ with and
without NUDGE in the VLASH framework~\cite{tang2025vlash}. Guidance adds an FK pass, an ESDF query, and a backward gradient pass at
each of the two enabled denoising steps, totaling only $2.46$~ms of
end-to-end overhead under \texttt{torch.compile}.
Each row of the table is the
median over $30$ timed iterations after $5$ warmup runs.
Per-stage timings use eager PyTorch with CUDA synchronization. Compiled
totals use \texttt{torch.compile}, whose CUDA graphs do not expose
per-stage timers.

%% file: content/07_conclusion.tex
\section{Discussion}

We presented NUDGE, an inference-time procedure that adds obstacle
avoidance to pretrained robot policies based on diffusion or flow matching
by injecting signed-distance-field gradients into the denoising loop.
NUDGE preserves task-manifold adherence and achieves 41\% collision-free
completion versus 16\% for the state-of-the-art baseline on 100 matched
single-obstacle LIBERO episodes. On a UR5, it achieves 75.6\% completion over 82 consecutive trials
with a moving obstacle. Guidance adds 2.46\,ms per action chunk on the UR5. Natural next steps are extending guidance to
articulated and deformable obstacles, modeling held-object geometry more
faithfully. Future work will also focus on providing a formal safety guarantee for robot policies.